\documentclass{article}
\usepackage{aaai}

\begin{document}

\title{Automatic Belief Revision in SNePS\thanks{The development of
    SNePS was supported in part by: the National Science Foundation
    under Grant IRI-8610517; the Defense Advanced Research Projects
    Agency under Contract F30602-87-C-0136 (monitored by the Rome Air
    Development Center) to the Calspan-UB Research Center; by the Air
    Force Systems Command, Rome Air Development Center, Griffiss Air
    Force Base, New York 13441-5700, and the Air Force Office of
    Scientific Research, Bolling AFB DC 20332 under Contract No.
    F30602-85-C-0008, which supported the Northeast Artificial
    Intelligence Consortium (NAIC); NASA under contract NAS 9-19335 to
    Amherst Systems, Inc.; and by ONR under contract N00014-98-C-0062
    to Apple Aid, Inc.  The development of automatic belief revision
    was supported in part by the U.S. Army CECOM through a contract
    with CACI Technologies.}\\
CSE Technical Report 2000-01
}

\author{Stuart C. Shapiro and Frances L. Johnson\\
Department of Computer Science and Engineering\\ 
and Center for Multisource Information Fusion\\
and Center for Cognitive Science\\
State University of New York at Buffalo\\
226 Bell Hall\\
Buffalo, NY 14260-2000, USA\\
\textsf{\{shapiro $\mid$ flj\}@cse.buffalo.edu}
}
\nocopyright
            
\maketitle

\begin{abstract}
\noindent 
SNePS is a logic- and network- based knowledge representation,
reasoning, and acting system, based on a monotonic, paraconsistent,
first-order term logic, with compositional intensional semantics.  It
has an ATMS-style facility for belief contraction, and an acting
component, including a well-defined syntax and semantics for primitive
and composite acts, as well as for ``rules'' that allow for acting in
support of reasoning and reasoning in support of acting.  SNePS has
been designed to support natural language competent cognitive agents.

When the current version of SNePS detects an explicit contradiction,
it interacts with the user, providing information that helps the user
decide what to remove from the knowledge base in order to remove the
contradiction.  The forthcoming SNePS~2.6 will also do automatic
belief contraction if the information in the knowledge base warrents
it.
\end{abstract}
\section{General Information}


\subsection{Platforms and Language}
The current version of SNePS is written in ANSI Common Lisp, and runs
on any platform that runs ANSI Common Lisp.

\subsection{Size}
To install SNePS, one needs about 10 Megabytes of disk space.
Once the installation is completed, this might be trimmed to about 5
Megabytes by compressing/deleting Lisp source files and/or
documentation.

\subsection{Additional Information}
More information may be found at the following URLs:
\begin{description}
\item[\texttt{http://www.cse.buffalo.edu/sneps/}] home page of the
  SNePS Research Group
\item[\texttt{http://www.cse.buffalo.edu/sneps/Bibliography/}] a
  bibliography of over 270 papers on SNePS
\item[\texttt{http://www.cse.buffalo.edu/sneps/Manuals/}] a
  repository of SNePS Manuals
\item[\texttt{ftp://ftp.cse.buffalo.edu/pub/sneps/}] the SNePS ftp site
\end{description}

At the time of writing this paper, the most current available version
of SNePS is SNePS~2.5.  This paper also describes SNePS~2.6, which
will be the next release.  The major new feature of SNePS~2.6 is that,
under certain circumstances, it will perform automatic belief revision
\cite{JohSha00}.

\section{Description of the System} 
SNePS is a logic- and network- based knowledge representation,
reasoning, and acting system  that has been developed over the course
of the last thirty years by the author and over 60 students and
colleagues \cite{ShaRap91a,ShaSIG99a}.  SNePS has been designed to
support natural language competent cognitive agents.  Its logic is
based on Relevance Logic \cite{Sha92}, a paraconsistent logic (in
which a contradiction does not imply anything whatsoever).  The basic
principles of SNePS are:
\begin{quote}
\begin{description}
\item[Propositional Semantic Network:] The only well-formed SNePS
  expressions are nodes.
\item[Term Logic:] Every well-formed SNePS expression is a term.
\item[Intensional Representation:] Every SNePS term represents
  (denotes) an intensional (mental) entity.
\item[Uniqueness Principle:] No two SNePS terms denote the same
  entity.
\end{description}
\end{quote}

The following, more system-oriented, description is a slight rewriting of the
Introduction to the SNePS~2.5 User's Manual \cite{ShaSIG99a}.

SNePS (the Semantic Network Processing System) is a system for
building, using, and retrieving information from propositional
semantic networks.  A semantic network, roughly speaking, is a labeled
directed graph in which nodes represent entities, arc labels represent
binary relations, and an arc labeled $R$ going from node $n$ to node
$m$ represents the fact that the entity represented by $n$ bears the
relation represented by $R$ to the entity represented by $m$.

SNePS is called a {\em propositional} semantic network because every
proposition represented in the network is represented by a node, not
by an arc.  Relations represented by arcs may be thought of as part of
the syntactic structure of the node they emanate from.  Whenever
information is added to the network, it is added in the form of a node
with arcs emanating from it to other nodes.

Each entity represented in the network is represented by a unique
node.  This is enforced by SNePS~2 in that whenever the user specifies
a node to be added to the network that would look exactly like one
already there, in the sense of having the same set of arcs going from
it to the same set of other nodes, SNePS~2 retrieves the old one
instead of building the new one.

The core of SNePS~2 is a system for building nodes in the network, retrieving nodes that have a certain
pattern of connectivity to other nodes, and performing certain housekeeping tasks, such as dumping a network
to a file or loading a network from a file.

SNIP, the SNePS Inference Package, interprets
certain nodes as representing reasoning rules, called {\em deduction
rules}.  SNIP supports a variety of specially designed propositional
connectives and quantifiers, and performs a kind of combined
forward/backward inference called {\em bi-directional} inference.

SNeBR, the SNePS Belief Revision system \cite{MarSha88}, recognizes
when a contradiction exists in the network, and interacts with the
user whenever it detects that the user is operating in a contradictory
belief space.  Under certain circumstances, SNePS~2.6 will perform
automatic belief contraction (\textit{see} \cite{JohSha00}).

SNeRE, The SNePS Rational Engine, is a package that allows for the
smooth incorporation of acting into SNePS-based agents, including
acting in the service of inference and \textit{vice versa}
\cite{KumSha94a,Kum94a}.

SNePSUL, the SNePS User Language, is the standard command language for
using SNePS.  It is a Lispish language, usually entered by the user at
the top-level SNePSUL read-eval-print loop, but it can also be called
from Lisp programs.

SNePSLOG is a logic programming interface to SNePS, and provides
direct access in a predicate logic notation to almost all the
facilities provided by SNePSUL.  A Tell/Ask interface allows SNePSLOG
expressions to be used within normal Lisp programs.

SNaLPS, the SNePS Natural Language Processing System, consists of a
morphological analyzer, a morphological synthesizer, and a Generalized
Augmented Transition Network (GATN) Grammar interpreter/compiler
\cite{Sha82}.  Using these facilities, one can write natural language
(and other) interfaces for SNePS.

XGinseng is an X Windows-based graphical editing and display
environment for SNePS networks.  XGinseng is the best environment to
use for preparing diagrams of SNePS networks for inclusion in papers.
It can also be used to build and edit SNePS networks.  (See the screen 
shot at \texttt{http://www.cse.buffalo.edu/sneps/screen.gif})


\section{Applying the System}


\subsection{Methodology}


Problems are encoded in SNePS logic, which can express any formula
expressible in first-order logic, but also contains features
specifically designed for a network-oriented KRR system for natural
language competence and commonsense reasoning \cite{Sha00a}, such as
set-oriented (instead of binary) connectives and numerical
quantifiers.  SNePS is an intensional term-logic, meaning, in part,
that propositions are denoted by functional terms, so propositions may
be arguments of propositions without the need for quotation, modal
logic, or leaving first-order logic \cite{Sha93a}.

To take full advantage of automatic belief revision in SNePS~2.6, the
user should give the system information about the sources of
information, should order the sources by credibility, and may provide
credibility ordering of the information directly.  The following shows
some information, source information, source credibility ordering
information, and direct information credibility ordering as it might
be given to the system using the SNePSLOG interface:
\begin{verbatim}
fun(learning).
~fun(spitting).

Source(Lisa, fun(learning)).
Source(Lisa, ~fun(spitting)).
Source(Bart, fun(spitting)).

Sgreater(Lisa,Marge).
Sgreater(Marge, Bart).
Sgreater(Bart,Homer).

Greater(fun(learning),~fun(spitting)).
\end{verbatim}
                                        

\subsection{Specifics}

\subsubsection{Significance of Being Logic-Based}                                        
SNePS is a knowledge representation, reasoning, and acting system.  We
believe that every knowledge representation and reasoning formalism
must have a well-defined syntax, a well-defined semantics, and a
well-defined inference mechanism, to implement reasoning, that is sound
with respect to the semantics.  Thus, we believe that every knowledge
representation and reasoning formalism is a logic, although it might
not be standard classical first-order predicate logic.

\subsubsection{Semantics}
SNePS is a paraconsistent, first-order term logic, with compositional
intensional semantics.  It is currently monotonic, although it has an
ATMS-style facility for belief contraction---removal of an assumption
from the belief space along with all derived beliefs that thereby
loose all their supports.  It also has an acting component, including
a well-defined syntax and semantics for primitive and composite acts,
as well as for ``rules'' that allow for acting in support of reasoning
and reasoning in support of acting.

\subsubsection{Importance}
SNePS has been designed, and continues to be developed so that a SNePS 
knowledge base can form the mind of a natural language competent
cognitive agent.  The features mentioned above are all in support of
this purpose.

SNePS is paraconsistent so that it can continue to reason, even while
containing contradictions, without a contradiction in one area of its
beliefs ``polluting'' its conclusions in an unrelated area.  This is
based on the fact that people, likewise, can have contradictory
beliefs without assenting to every question.

SNePS is first-order to make use of well-known first-order inference
techniques.  Nevertheless, the ``end-user'' uses a language that
includes only the individual symbols of SNePS, so that language looks
higher-order. (\textit{See} \cite{ShaMckMarMor81}.)

SNePS is a term logic, meaning that every well-formed expression is a
term in the language---there are no sentential-level expressions.  For
example, since function symbols can take functional terms as
arguments, and propositions are considered to be first-class
individuals in the domain, denoted by terms, propositions can take
propositions as arguments without the need for quotation, modal logic,
or leaving first-order logic.  (\textit{See} \cite{Sha93a}.)  This was 
illustrated near the end of the subsection on methodology by the
SNePSLOG expression \texttt{Source(Lisa, fun(learning))}, where the
functional term \texttt{fun(learning)} denotes the proposition that
learning is fun, and the functional term \texttt{Source(Lisa,
  fun(learning))} denotes the proposition that Lisa is the source of
the information that learning is fun.  Again, this is based on the
fact that people talk about propositions, treating them as individuals 
in the domain of discourse.

SNePS has a compositional semantics for the usual reason that a single 
term can be included as an argument in multiple functional terms while 
maintaining a single denotation.

Every SNePS term denotes an intensional, or mental, entity, because
SNePS knowledge bases are intended to serve as minds of cognitive
agents, and no two entities that are conceptually different are
identical.  Even the equation $2+3=5$ is informative only if $2+3$ and 
$5$ denote conceptually distinct entities.  (\textit{See}
\cite{MaiSha82,ShaRap85}.)

The current version of SNePS is monotonic, although previous versions
contained default rules, and we intend to reintroduce them in a future 
version.

Although SNePS is paraconsistent, we believe that when a contradiction
becomes explicit, the user should be afforded the opportunity of
removing the contradiction by removing some hypothesis that underlies
it.  An explicit contradiction, the presence of both some proposition
\texttt{P} and its negation, \verb|~P|, in the belief space, is easily
recognized by the system because, in accordance with the Uniqueness
Principle, the data structure representing \texttt{P} is directly
pointed to by the negation operator in the data structure representing
\verb|~P|.  In SNePS~2.6, when the choice of which hypotheses to
remove is ``obvious'', the system will do so automatically, and notify
the user of the hypothesis removed and of the other beliefs that are
no longer in the belief space because they are no longer supported.

Since SNePS is designed for cognitive agents, it is important for
acting and reasoning to be integrated.  For example, if \texttt{light1}
denotes some traffic light, \texttt{street1} the street where that
traffic light is, \texttt{green({\itshape x})} the proposition that
\texttt{\itshape x} is green, and \texttt{cross({\itshape x})} the act
of walking across \texttt{\itshape x}, then
\texttt{whendo(green(light1), cross(street1))} denotes the proposition
that when the agent comes to believe that the traffic light is green,
it should cross the street.  If, moreover, \texttt{lookat({\itshape
    x})} denotes the act of looking at \texttt{\itshape x}, then
\texttt{ifdo(green(light1), lookat(light1))} denotes the proposition
that if the agent wants to know whether to believe that the traffic
light is green, it should look at it.\\[1.5ex]

\subsubsection{Influence}
SNePS and its immediate predecessors (\textit{see} \cite{ShaRap91a})
have been influential in the fields of artificial intelligence,
knowlege bases, and deductive databases.  It was the first
network-based knowledge representation system to clearly distinguish
``system relations,'' represented by arcs from ``conceptual
relations'' represented by nodes, the first network-based knowledge
representation system to have a way of representing all of first-order
logic, the first reasoning system to be able to reason with recursive
rules without getting into an infinite loop, the first knowledge
representation system to be fully and exclusively intensional, the
first knowledge representation system to represent propositions about
propositions without the need for quotation or modal operators, and
the first reasoning system to include an assumption-based truth
maintenance system.  We believe that SNePS is the first system to
integrate reasoning and acting in a syntactically and semantically
clean way.  In our current work on automatic belief revision, we are
setting out to explicate how integrity constraints and belief revision
postulates can be applied to and implemented in deductively open
belief spaces.

\subsection{Users and Useability} 
The potential users of SNePS should have a familiarity with logic at
least at the level of students of an introductory logic course. A
tutorial may be accessed from the SNePS Research Group web pages.
SNePS has not been designed to apply to any specific area of
application, so it is flexible enough to handle a wide variety of
areas, but all the domain knowledge must be supplied by the user.  In
that sense, SNePS is like an expert system shell or logic programming
language.  However, it is different in having been designed to build
cognitive agents, so the user is an informant rather than a
programmer, and it is sometimes difficult even for the one user who
has supplied all the domain knowledge to predict the conclusions SNePS
will reach.  Nevertheless, SNePS has been, and is being used by people
outside the SNePS Research Group.

\section{Evaluating the System}


\subsection{Benchmarks}
SNePS comes with a suite of demonstration problems and applications
that can be used to familiarize oneself with how to use it, and may be 
used for comparison with other systems.  Demonstrations that were
taken from other sources include The Jobs Puzzle from
\cite[Chapter~3.2]{WosOveLusBoy84}, Schubert's steamroller problem
(\textit{see} \cite{Sti85}), and a database management system example
from \cite{Dat81}.

SNePS is moderately user friendly.  It is ``academic'' software,
rather than a commercial product, and this aspect could be somewhat
improved.

\subsection{Comparison} 

Schubert's steamroller problem was run on a 1993 version of SNePS
\cite{Cho93}, and the results compared with those reported in
\cite{Sti85}.  The SNePS version produced fewer unifications and was
faster than most unsorted logic solutions, but was outperformed by
sorted logic solutions.  The current version of SNePS is much faster
on this problem than the 1993 version, partially due to improvements
in SNePS, and partially due to faster, bigger computers.

The SNePS representation of the Jobs Puzzle is much simpler and closer 
to the English version of the puzzle than the clause form
representation presented in \cite[p. 58ff]{WosOveLusBoy84}.

\subsection{Problem Size} 
Currently, SNePS can handle knowledge bases on the order of about
1,000 SNePS terms.


SNePS~2.5 is more than a prototype system.  It is useable for
research and experimentation.  However, we would not claim it to be ready
for full-scale applications.  Before it would be ready for large-sized 
problems, it would need a concentrated effort of profiling and
optimization.

\bibliography{/u0/faculty/shapiro/Util/Latex/bibliography}
\bibliographystyle{aaai}

\end{document}